# Title Page

## Title

Multi-omics data integration for early diagnosis of hepatocellular carcinoma (HCC) using machine learning

## Author Information

Annette Spooner[1]*, Mohammad Karimi Moridani[2], Azadeh Safarchi[6], Salim Maher[4,5], Fatemeh Vafaee[2,3], Amany Zekry[4,5], Arcot Sowmya[1]

[1] School of Computer Science and Engineering, University of New South Wales, Sydney, Australia.
[2] School of Biotechnology and Biomolecular Sciences, University of New South Wales, Sydney, Australia.
[3] UNSW Data Science Hub, University of New South Wales, Sydney, New South Wales, Australia
[4] St George and Sutherland Clinical Campuses, University of New South Wales, Sydney, Australia
[5] Department of Gastroenterology and Hepatology, St George Hospital, Sydney, Australia
[6] Health and Biosecurity, Commonwealth Scientific and Industrial Research Organisation, Sydney, Australia.

**ORCIDS:**
Annette Spooner: 0000-0002-5705-0602
Mohammad Karimi Moridani: 0000-0003-0793-3797
Azadeh Safarchi: 0000-0002-9112-939X
Salim Maher: 0000-0002-2579-5925
Fatemeh Vafaee: 0000-0002-7521-2417
Amany Zekry: 0000-0002-5675-1810
Arcot Sowmya: 0000-0001-9236-5063

**\*Corresponding author:** Annette Spooner (a.spooner@unsw.edu.au)

## Abstract

The complementary information found in different modalities of patient data can aid in more accurate modelling of a patient's disease state and a better understanding of the underlying biological processes of a disease. However, the analysis of multi-modal, multi-omics data presents many challenges, including high dimensionality and varying size, statistical distribution, scale and signal strength between modalities.

In this work we compare the performance of a variety of ensemble machine learning algorithms that are capable of late integration of multi-class data from different modalities. The ensemble methods and their variations tested were i) a voting ensemble, with hard and soft vote, ii) a meta learner, iii) a multi-modal Adaboost model using a hard vote, a soft vote and a meta learner to integrate the modalities on each boosting round, the PB-MVBoost model and a novel application of a mixture of experts model. These were compared to simple concatenation as a baseline.

We examine these methods using data from an in-house study on hepatocellular carcinoma (HCC), along with four validation datasets on studies from breast cancer and irritable bowel disease (IBD). Using the area under the receiver operating curve as a measure of performance we develop models that achieve a performance value of up to 0.85 and find that two boosted methods, PB-MVBoost and Adaboost with a soft vote were the overall best performing models.

We also examine the stability of features selected, and the size of the clinical signature determined. Finally, we provide recommendations for the integration of multi-modal multi-class data.

## Statements and Declarations

### Conflicts of interest/Competing interests

Author Fatemeh Vafaee is a member of the editorial board of Artificial Intelligence Review.

### Funding

This work was funded by a grant from the Medical Research Future Fund (MRFF), grant id 2008996.

### Author contributions

A.Sp. designed the machine learning experiments, carried out the experiments and prepared the manuscript. A.So. and F.V. provided expert guidance and reviewed the manuscript. A.Z. provided medical guidance, provided data and reviewed the manuscript. A.Sa. assisted with the validation datasets and reviewed the manuscript. S.M. provided a description of the in-house dataset. M.K.M. reviewed the manuscript.

### Data Availability

The HCC dataset is available from the authors upon reasonable request. Data availability for the other datasets can be found in the relevant papers – references are given in Table 1.

### Code Availability

The code developed in this project will soon be available on Github.

# Multi-omics data integration for early diagnosis of hepatocellular carcinoma (HCC) using machine learning

## 1. Introduction

Modelling complex biological systems has the potential to expand our understanding of many challenging diseases. Data encompassing entire biological systems, known as 'omics data, can provide potentially complementary information, giving a more accurate picture of a patient's disease state. Multi-omics data integration is the process of combining two or more 'omics datasets in order to gain a better understanding of the mechanisms of biological processes and the complex interactions between biological systems (Krassowski et al. 2020) (Chicco, Cumbo, and Angione 2023). Multi-omics data integration can improve the accuracy of clinical outcome prediction, provide novel insights into mechanisms underlying complex diseases, aid in subtyping diseases and stratifying patient cohorts, discover new biomarkers or identify potential therapeutic targets (Chicco, Cumbo, and Angione 2023) (Picard et al. 2021) (Dickinson et al. 2022) (Kreitmaier, Katsoula, and Zeggini 2023).

However, the analysis of multi-omics data presents many challenges. 'Omics datasets typically contain a large number of features and only a small number of samples, because of the high cost of clinical data collection and the limited number of study participants (Tabakhi et al. 2023). If not addressed, this leads to overfitting and therefore biased results in most machine learning algorithms. In addition, these datasets often contain many irrelevant and redundant features, misleading the algorithm and increasing computational complexity. The different 'omics datasets are often heterogeneous, having been collected using different technologies, and can vary greatly in size, statistical distribution, scale and signal strength (Santiago-Rodriguez and Hollister 2021). They may be imbalanced in terms of the number of features or the composition of the classes. In addition, they may have missing values or suffer from batch effects. Finally, multi-omics data integration must provide stable and interpretable results if it is to be trusted by clinicians.

Data integration strategies are often categorised as early, intermediate or late (Serra, Galdi, and Tagliaferri 2018). Early integration, also known as feature-level fusion, simply concatenates all of the 'omics data into a single dataset and trains a classifier on it directly. This exacerbates the problems of high-dimensionality, noise and highly correlated features (Chicco, Cumbo, and Angione 2023). Intermediate integration transforms the omics datasets into a common representation space prior to modelling (Wang et al. 2022). It can capture the relationships between the various modalities, but clinical interpretation of the results is difficult. Late integration, also known as decision-level fusion, trains a model on each 'omics dataset independently and the results are aggregated to give a final prediction. Techniques for aggregating the results may include majority vote, weighted majority vote, sum of the probabilities in each class and meta learning (Galar et al. 2011). Late integration is illustrated in Figure 1.

Late integration or decision-level fusion is well placed to overcome many of the challenges of multi-omics data integration. By training a separate model on each omics modality, the curse of dimensionality is reduced, meaning overfitting is less likely to occur and computational complexity is also reduced. The approach is flexible as different machine learning models can be trained on each modality, taking advantage of the best performing model in each case and also addressing the problems of heterogeneity and feature imbalance between modalities (Tabakhi et al. 2023). Pre-processing steps such as filtering, imputation of missing values, normalisation and feature selection, can also be tailored to individual modalities.

Despite the many advantages of late integration strategies in overcoming the challenges of multi-omics data integration, there is a paucity of literature examining these methods. Previous studies have examined the use of machine learning for multi-modal data integration in various clinical settings including oncology (Raufaste-Cazavieille, Santiago, and Droit 2022) (Acharya and Mukhopadhyay 2024), breast cancer specifically (Sammut et al. 2022), autoimmune disease (Martorell-Marugán et al. 2023), inflammatory bowel disease (Gardiner et al. 2022) and in studies of type 2 diabetes, osteoarthritis, Alzheimer's disease, and systemic lupus erythematosus (Kreitmaier, Katsoula, and Zeggini 2023). However, none of these studies focused on the advantages of late integration.

Our work aims to fill this gap by benchmarking late integration strategies using multi-modal ensemble machine learning. Multi-modal machine learning is the task of learning from multiple modalities of the data to exploit complementary information and improve learning performance (Kline et al. 2022). Ensemble machine learning is a technique that combines multiple weak learners to form a strong learner that is more accurate than its base learners (Zhao et al. 2017). Multi-modal ensemble learning, therefore, combines these two techniques and can integrate and learn from multiple sources of high-dimensional 'omics data.

In this work we benchmarked eight ensemble machine learning methods, including multiple variations, for the late integration of multi-modal, multi-class data, and compared these methods to a simple concatenation of the data. We applied these methods to an in-house dataset containing data collected from patients with liver disease (Behary et al. 2021), and validated them on data from four publicly available multi-omics datasets. We used existing methods and also improved some methods to enhance their predictive accuracy. We examined the predictive accuracy of these methods as well as the stability of the features they selected and compared the results of the multi-modal methods with those of the individual modalities. For each dataset, we also determined an optimal subset of modalities which performed as well as the full set, thereby permitting patient diagnosis using fewer tests.

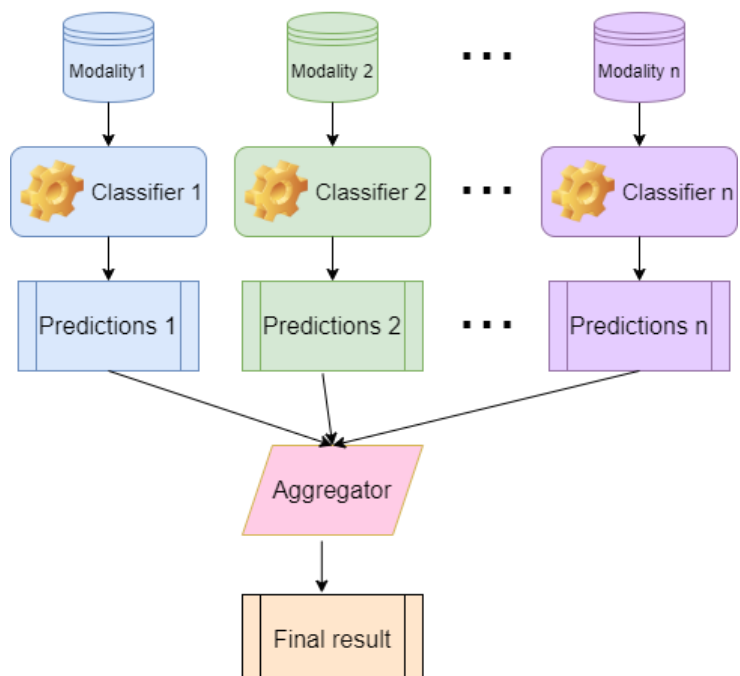

*Figure 1 – Late integration.*

## 2. Methods

To demonstrate the benefits of ensemble machine learning for multi-omics data integration, we benchmarked five late-integration methods that are capable of multi-modal, multi-class learning, and tested their predictive accuracy as well as their suitability for knowledge discovery, measured by their ability to select a stable and interpretable set of predictive features. The machine learning methods tested include: a) a simple voting ensemble, that aggregates the results from the different modalities using either a hard or soft vote; b) meta-learning, an algorithm that learns from the outputs of other machine learning algorithms; c) an enhanced multi-modal version of the well-known boosting algorithm Adaboost (Freund and Schapire 1995); d) a multi-view boosting method known as PB-MVBoost (Goyal et al. 2019), which takes into account both the accuracy and the diversity of the classifiers trained on each view and e) a mixture of experts model that trains a different model for each class and integrates these with a gating function. We compared these methods to a single classifier trained on the concatenated data as a baseline.

### 2.1. Data

The proposed multi-modal, multi-class machine learning methods were applied to an in-house dataset containing data collected from patients with liver disease (Behary et al. 2021), and validated on data from four publicly available multi-omics datasets, two containing data from patients with inflammatory bowel disease (IBD) and two from patients with breast cancer, to differentiate between different stages and types of disease. The characteristics of these datasets are summarised in Table 1, where references are given.

The in-house dataset (Behary et al. 2021) consists of data collected from prospectively enrolled patients to investigate the effect of the gut-microbiota's composition and function on the development of liver disease and primary hepatocellular carcinoma (HCC). Adult study subjects with liver cirrhosis and/or HCC were enrolled from two liver centres in Sydney Australia, as were healthy adult volunteers to function as controls. Baseline demographic and clinical data were collected at time of enrolment, with blood samples collected for multi-omics data and oral/faecal samples collected for metagenomic sequencing of oral/gut microbiota.

Study subjects were subsequently divided into one of four classes: healthy controls (CON); individuals with liver cirrhosis secondary to metabolic-associated fatty liver disease (MAFLD-cirrhosis) (CIR); individuals with HCC secondary to MAFLD-cirrhosis (LN); individuals with HCC secondary to viral hepatitis (LX). Of the 122 study subjects, the number of samples common to all modalities was 106, and samples were approximately evenly distributed across classes, as shown in Table 1. Seven modalities of data were available, and these are also listed in Table 1. For the stool and oral microbiome, two microbial levers of genera and species were used separately in the models.

| DB Name | Reference | Patient categories (Abbreviation) | No. Samples | Omics modalities (Abbreviation) | No. Features * |
|---|---|---|---|---|---|
| HCC-Genus HCC-Species | *Private unpublished data* | Healthy Controls (CON) MAFLD-cirrhosis (CIR) MAFLD-related HCC (LN) Viral HCC (LX) | 28 28 25 25 | Clinical (CLIN), Cytokine (CYT), Pathology results (PATH), Metabolomic (MET) Lipoprotein (LIP), Oral Microbiome-Genus (OG), Oral Microbiome-Species (OS), Stool Microbiome-Genus (SG), Stool Microbiome-Species (SS) | 14 28 48 1046 112 243 583 282 721 |
| IBD-1 | Mehta et al. (2023) (Mehta et al. 2023) Nature medicine DOI: 10.1038/s41591-023-02217-7 | Crohn's disease (CD) Ulcerative Colitis (UC) Non-IBD (non-IBD) | 50 28 20 | Metagenomics (MTG) Metabolomics (MTB) Metatranscriptomis (MTX) Viromics (VIR) | 934 81,496 83,227 262 |
| IBD-2 | Franzosa et al. (2019) (Franzosa et al. 2019) Nature Microbiology DOI: 10.1038/s41564-018-0306-4 | Crohn's disease (CD) Ulcerative Colitis (UC) Control | 88 76 56 | Clinical (CLIN) Metabolites (METAB) Microbiome (MICROB) | 8 8850 204 |
| Breast-1 | Sammut et al. (2022) (Sammut et al. 2022) Nature DOI: 10.1038/s41586-021-04278-5 | RCB-1 RCB-II RCB-III pCR | 24 59 27 40 | Clinicopathological (CLIN) Digital pathology (PATH) RNA sequencing (RNA) DNA sequencing (DNA) | 24 8 57,903 31 |
| Breast-2 | Krug et al. (2020) (Krug et al. 2020) Cell DOI: 10.1016/j.cell.2020.10.036 | Basal-like (Basal) HER2-enriched (Her2) Luminal A (LumA) Luminal B (LumB) Normal-like (Normal) | 29 14 57 17 5 | Clinical (CLIN) mRNA (MRNA) Proteome (PROT) | 28 23123 9932 |

*Table 1. Summary of the characteristics of the datasets used in the study, showing patient categories and number of samples in each, plus modalities and the number of features in each. The order of the number of samples and number of features is consistent with the list of patient categories and modalities respectively.*

## 2.2. Data Pre-processing

The data processing pipeline is shown in **Error! Reference source not found.Error! Reference source not found.Error! Reference source not found.**. Special consideration was given to memory management because of the large data file sizes, resulting in a 2-step process to read the files into memory. The first step was to read the raw files one at a time, apply filtering to identify potentially relevant features, as opposed to those with no predictive ability, record their names, then release the memory used. In the second step, only the relevant features of all files were read in, reducing the amount of memory required.

Prior to modelling, filtering was performed to identify potentially relevant features and reduce the dimensionality of the data. This was also a multi-step process, that could be tailored to each dataset and depended on the characteristics of the dataset. The following command line options could be chosen:
1. Features with more than 50% missing values and more than 90% zero values were eliminated.
2. Of the remaining features, one feature from each pair of correlated features was eliminated.

3. If the dimensionality of the dataset was still very large (i.e. if the ratio of the number of features to the number of samples was greater than 10), then only the 500 features with the highest variance were retained.

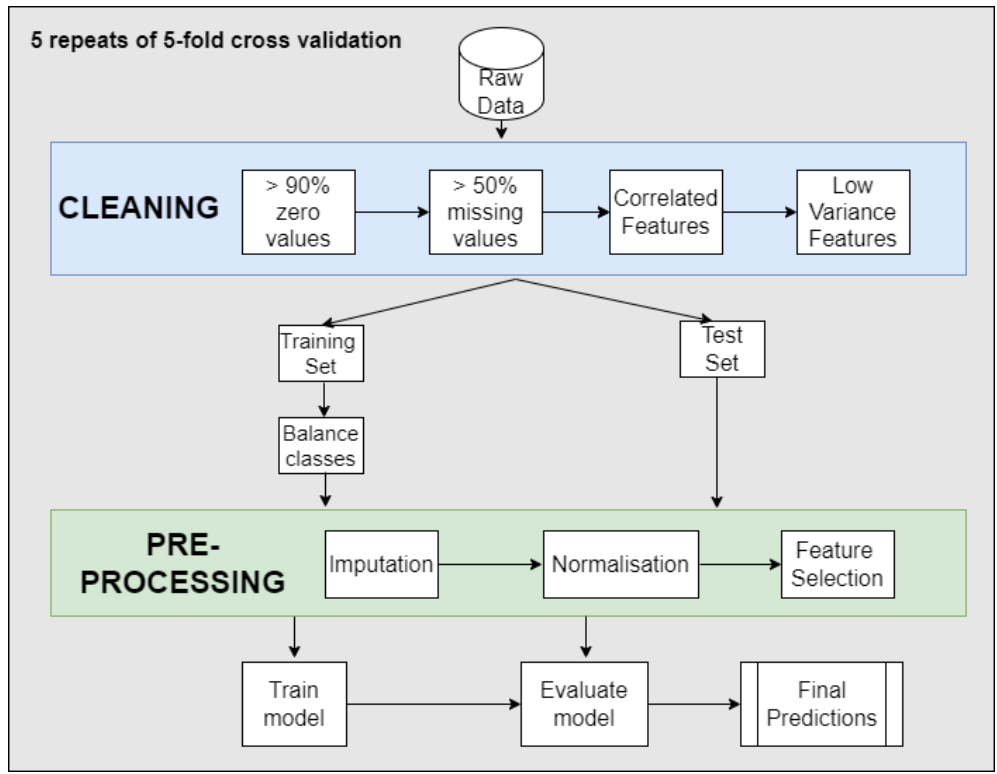

*Figure 2. Data processing pipeline.*

The following pre-processing steps were applied to the data during model building:
1. **Balancing:** for imbalanced datasets, classes were balanced using the Synthetic Minority Oversampling Technique (SMOTE) (Chawla, N.V., Bowyer, K.W., Hall, L.O. 2002). SMOTE was applied to the training set only, during model building, following the method of (Rachmatullah 2022), balancing one class at a time against the majority class.
2. **Imputation**: missing values were imputed using Multiple Imputation by Chained Equations (MICE) (van Buuren and Oudshoorn 2007). However, if the size of the dataset caused the computation time of MICE to increase unacceptably, then kNN, a single imputation method, was employed instead.
3. **Normalisation:** Counts per Million (CPM) normalisation, followed by a log transformation was applied to the RNA sequence and DNA data. Standardisation was applied to all other data.
4. **Feature selection:** Feature selection was performed using the Boruta feature selection algorithm (Kursa, Jankowski, and Rudnicki 2010), with the Gradient Boosting Machine (Freund and Schapire 1995) as the underlying learner.

Boruta (Kursa, Jankowski, and Rudnicki 2010) is a model-agnostic algorithm, which can wrap around any base learner that provides feature importance scores. It creates a set of shadow features, which are randomly shuffled copies of the original features, then repeatedly trains a random forest on the combined set of features. Any features that achieve a lower importance score than the highest-scoring shadow feature are considered irrelevant. In this way Boruta naturally generates a feature selection threshold, which is based on p-values.

Boruta is known as an "all-relevant" method, as it identifies all features relevant to the target variable, including those that are correlated. This is in contrast to the more common "minimal-optimal" feature selection methods, which identify a small subset of features that maximise predictive accuracy. The "all-relevant" method is advantageous when the aim of model development is knowledge discovery.

### 2.3. Individual modalities

In order to see the benefits of data integration, a classifier was first trained on each of the individual modalities, without performing data integration. The Gradient Boosting Machine (GBM) classifier was chosen because of its superior performance in previous experiments. The aim of these experiments was not only to set a baseline for comparison with the integration techniques, but also to observe which modalities gave the best predictive performance.

### 2.4. Data Integration Techniques

Machine learning methods capable of integrating multiple modalities of data using the late integration strategy were examined and compared in this study. To ensure a fair comparison, all methods used the gradient boosting machine (GBM) (Freund and Schapire 1995), a multi-class classifier, as their underlying classifier. However, it should be noted that the methods investigated allow different underlying classifiers to be trained for each modality. The methods, which were developed using custom code, are summarised in Table 2 and the integration methods are described graphically in Figure 3.

| Integration Method | Results Aggregator | Abbreviation | Brief Description |
|---|---|---|---|
| Concatenation | | CONCAT | All modalities are joined into a single dataset and a classification model is trained on this dataset. |
| Voting Ensemble | Hard vote | ENS-H | A classification model is trained on each modality and the predictions are combined using majority vote. |
| Voting Ensemble | Soft vote | ENS-S | A classification model is trained on each modality and the predictions are combined by adding the class probabilities. |
| Meta Learner | Meta learner | ML | A classification model is trained on each modality and the predictions are combined by training a learner (in this case a random forest) on the results of the individual models. |
| Adaboost | Hard vote | ADA-H | A multi-modal Adaboost model is trained using all modalities. In each boosting round the predictions are combined using a majority vote. |
| Adaboost | Soft vote | ADA-S | A multi-modal Adaboost model is trained using all modalities. In each boosting round the predictions are combined using the sum of probabilities |
| Adaboost | Meta learner | ADA-M | A multi-modal Adaboost model is trained using all modalities. In each boosting round the predictions are combined using a meta learner. |
| PB-MVBoost | Weighted vote | PBMV | A classification model is trained on each modality and weights are learned for each modality and each classifier. Results are combined using a weighted vote. |
| Mixture of Experts using voting ensemble | Gating function | MOE-COMBN | A multi-modal voting ensemble classifier is trained for each class and a gating function selects the most confident result for each sample. |

*Table 2 Data Integration methods benchmarked in- this study.*

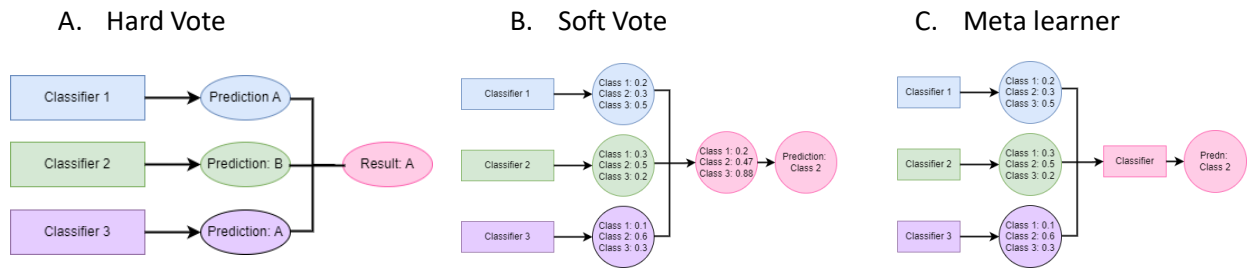

*Figure 3 Aggregation Techniques*

### 2.4.1. Concatenation

This method, also known as feature-level fusion (Tabakhi et al. 2023), combines the data from all modalities into a single vector and trains a single multi-class classifier on the concatenated data. This method was included as a baseline comparison for the other, more advanced methods.

### 2.4.2. Voting Ensemble

A voting ensemble trains a classifier on each modality and aggregates the outputs of the individual classifiers using a hard vote or a soft vote, illustrated in Figure 3. In a hard vote, or majority vote, the class that is predicted by the greatest number of modalities is the final prediction. In the case of a tie, the first predicted class is selected. In a soft vote, or weighted vote, the probability scores from each modality for each class are averaged. The class with the highest probability score is the predicted class.

### 2.4.1. Meta Learner

Meta-learning is a general term for machine learning algorithms that learn how to combine the outputs of other algorithms to maximise predictive accuracy. In the case of multi-modal data, a different base learner can be trained on each modality in parallel, and the meta-learner trained on the outputs of those base learners. The advantage of meta-learning is that it can produce reliable results with only a relatively small number of examples (Rafiei et al., n.d.). A recent survey gives many examples of the use of meta-learning in healthcare, in areas such as clinical risk prediction, disease diagnosis and drug interaction detection (Rafiei et al., n.d.).

In this study, a meta learner, illustrated in Figure 3C, was constructed using the Gradient Boosting Machine (GBM) as the base classifier and a random forest as the meta learner.

### 2.4.2. Multi-modal Adaboost

Boosting is an ensemble technique that trains a series of weak classifiers sequentially, such that each classifier learns from the mistakes of its predecessors (Schapire 1990). Ultimately these weak classifiers are combined to form a strong classifier.

Adaboost, or Adaptive Boosting (Freund and Schapire 1995), was one of the first boosting algorithms to be proposed. On each iteration of the Adaboost algorithm, the samples that were misclassified in the last iteration are given increased weight, forcing the algorithm to focus on the more difficult to classify samples, with the aim of correcting the errors made in the last iteration. The final model is a weighted linear sum of all the models in the ensemble.

Adaboost was initially developed for binary classification, but a multi-class version of Adaboost was introduced by Zhu et al. (Zhu et al. 2009). More recently, multi-view versions of Adaboost have been proposed, which can also be used to model multi-modal data. Xu and Sun developed a multi-view Adaboost algorithm, but it was limited to two views (Xu and Sun 2010). Xiao and Guo extended the multi-view Adaboost framework to an arbitrary number of views, in the context of multilingual subjectivity analysis (Xiao and Guo 2012). They developed two approaches, each of which trains a learner separately on each view, then combines the results of the single view classifiers, on each iteration, either using a hard majority vote or a linear weighted combination of the outputs of the single-view classifiers.

Here a multi-modal version of Adaboost was implemented following the method of Xiao and Guo (Xiao and Guo 2012), with some novel modifications. On each round of the boosting process, a classifier was trained independently on each modality. The results from these classifiers were then aggregated to give a final decision using one of three methods – a hard vote, a soft vote or a meta learner trained on the results of the independent classifiers.

While in the original Adaboost it is a simple matter to identify the misclassified samples to calculate the classification error rate, in multi-modal Adaboost, all modalities must be taken into account when calculating the error rate. Here, a sample is considered to be correctly classified only if it is classified with high confidence. If the final decision is made by a hard vote or meta learner, a sample is classified with high confidence if at least half of the modalities agree on its classification. The soft vote gives a final probability for each class and a sample is classified with high confidence if the highest probability is at least double that of the next highest probability. If these conditions are not met, then the sample is considered to be misclassified.

The optimal number of boosting steps within each iteration was chosen empirically, based on tests run using different numbers of boosting steps, and was determined to be 20. Beyond this value no added benefit was achieved.

### 2.4.3. PB-MVBoost

PB-MVBoost (Goyal et al. 2019) is a multi-view ensemble method, based on Adaboost, that aims to balance the accuracy of the classifiers trained on each view and the diversity of their outputs by learning two sets of weights – weights over the classifiers and weights over the views. It then combines the results of each classifier using a weighted vote, learning the weights by minimising an upper-bound on the error of the majority vote. Fadnavis et al. used PB-MVBoost in a novel framework to distinguish healthy controls from those in the early stages of Huntington's disease (Fadnavis, Polosecki, and Garyfallidis 2021).

PB-MVBoost was implemented in R, directly following the author's Python implementation, which is available on Github.

### 2.4.1. Mixture of Experts

A mixture of experts model (Yuksel, Wilson, and Gader 2012) divides a complex machine learning task into multiple sub-tasks, based on domain knowledge, and trains a model on each sub-task. Each of these models focuses solely on its specific sub-task, becoming an expert in that sub-space. A gating function learns which expert to trust and selects the best expert to predict each sample. Minoura et al. (Minoura et al. 2021) developed a model for integrated analysis of single-cell multi-omics data using a mixture of experts model.

Here a novel adaptation of this method was developed: a separate model was trained for each class in the multi-class setting, using a one-vs-rest approach i.e. each expert was a binary classifier, distinguishing its own class from all other classes combined. The experts were trained in parallel and the subsequent gating function operated using the following rules:
- Each expert can only predict its own class or "REST".
- If an expert predicts its own class, and it is the only one to do so, then its prediction is accepted as correct.
- If more than one expert predicts its own class, then the prediction of the expert that predicts with the highest confidence (probability) is accepted as correct.

- If no experts predict their own class (i.e. all predict "REST"), then the sample is classified as unknown. This indicates which samples are difficult to predict.

## 2.5. Incremental Method

From a clinical perspective, the ability to make an accurate prediction from a smaller subset of modalities means that fewer tests are required, making diagnosis simpler, less expensive and possibly less invasive for the patient. With this in mind, we designed an incremental model which determines the subset of the modalities that gives maximum predictive performance.

The incremental model adds (or removes) one modality at a time to (or from) the model. Here we included all modalities at the start, and eliminated one modality at a time, comparing the performance after each elimination to the previous best performance. On each iteration, the modality that caused the largest drop in performance was removed, and subsequent tests were carried out on the reduced modality set. The modalities were integrated using a soft voting ensemble and the metric used for comparison was the F1 score. A small margin of error was allowed in the performance comparison.

## 2.6. Feature Selection and Calculation of Feature Importance Scores

Feature selection identifies the features that are most relevant to the model outcome (Li and Wang 2018). These features can then be examined as potential biomarkers or may provide useful insights into the underlying mechanisms of disease.

When multiple classifiers are combined to give a final result, the final feature importance scores must be calculated in a meaningful way. For most models, the final set of selected features consists of those selected by each individual classifier, normalised and scaled to a range of [0,1] for comparison.

Some models required additional consideration. The features provided as input to the meta learner are the results of the base classifiers applied to each modality. Therefore, the feature importance scores generated by the meta learner give an indication of the relative importance of each modality.

The calculation of feature importance scores for Adaboost and PB-MVBoost was more complex. For each fold of data to which the model was applied, multiple boosting iterations were run. Each iteration produced a model weight and a set of feature importance scores for each modality. The final feature importance scores for each modality in each fold were calculated as the sum of the raw scores multiplied by the model weight, divided by the sum of the weights.

The Mixture of Experts model selects a set of features for each expert i.e. for each class. The fact that these feature sets can differ shows clearly that different features can influence the model for different classes.

In addition, for all models, only those features that were selected in 75% of cross-validation iterations and that had a normalised score of 0.5 or higher were included in the final set of selected features.

## 2.7. Feature Selection Stability

Stability of feature selection can be defined as the reproducibility of the features selected, when the method is applied to different samples of the data (Turney 1995). Various measures of stability have been proposed, each of which has its own strengths and limitations. The relative weighted consistency index, proposed by Somol and Novovičová (Somol and Novovičová 2010), has been chosen here because of its ability to measure the stability of sets of features of different lengths, such as those selected from different data samples when running repeated experiments, and because it does not over-emphasise low-frequency features.

## 2.8. Experimental Framework and Model Evaluation

The methods being assessed were evaluated their predictive performance as well as the stability of the set of features they selected. All models were evaluated using 5 repeats of 5-fold cross-validation. The evaluation metrics shown in Table 33 were calculated for each fold and averaged over folds and classes to give a final result (M and M.N 2015).

| Metric | Description | Formula |
|---|---|---|
| Accuracy (acc) | The ratio of correct predictions to the total number of samples. | $\frac{tp + tn}{tp + fp + tn + fn}$ |
| Sensitivity (sens) | The fraction of positive samples that are correctly classified | $\frac{tp}{tp + fn}$ |
| Specificity (spec) | The fraction of negative samples that are correctly classified | $\frac{tn}{tn + fp}$ |
| Precision (p) | The fraction of samples identified as positive that were correctly classified | $\frac{tp}{tp + fp}$ |
| Recall (r) | As for sensitivity | $\frac{tp}{tp + fn}$ |
| F1 Measure (f1) | The harmonic mean between precision and recall | $\frac{2 * p * r}{p + r}$ |
| AUC (auc) | Area under the Receiver Operating Curve: a measure of the classifier's ability to distinguish between classes | |

*Table 3. Evaluation metrics calculated on all models*

All experiments were run on the Katana high performance computing cluster at the University of NSW ("Katana" 2010). All code was written in R (R Core Team 2019). The R package *mlr* (Bischl et al. 2016) was used as a framework to implement the machine learning experiments and the R package *Future* was used to implement parallel processing (Bengtsson 2021).

Tests of statistical significance were applied to each group of experiments, using the corrected resampled paired t-test, proposed by Nadeau and Bengio (Nadeau and Bengio 2003). This test takes into account the fact that the Type I error is inflated when applying a standard t-test on results from a repeated k-fold cross validation because the results are not independent and corrects for this.

# 3. Results

## 3.1. Individual modalities

The performance of the individual modalities in each dataset is shown in the boxplots in Supplementary Figures S1-S5 and in Supplementary Table S1. The values shown represent the average across all iterations of cross-validation.

In the HCC dataset, the best performing modality was the Cytokine data, with an AUC of 0.77 and an F1 score of 0.64 (±0.15). The metabolomic data also performed well, with an AUC of 0.74 and an F1 score of 0.6 (±0.15). In two other datasets, IBD1 and IBD2, the metabolomic data also proved to be the most predictive with AUC scores of 0.56 and 0.76 respectively and F1 scores of 0.39 (±0.17) and 0.68 (±0.06) respectively. In the breast cancer datasets, the clinical data was the most predictive in the Breast1 dataset, with an AUC of 0.8 and an F1 score of 0.65 (±0.2), and the proteomic data was the most predictive in the Breast2 dataset, with an AUC of 0.74 and an F1 score of 0.57 (±0.36).

## 3.2. Multi-modal Data Integration

Each of the muti-modal data integration strategies was applied to each dataset and the results of these experiments are shown in the boxplots in Figure 4 for the HCC (Genus) dataset, and in Supplementary Figures S6-S11 and Supplementary Table S2 for the remaining datasets. The PB-MVBoost method was either the best or second-best performing model in every dataset, achieving an AUROC score of 0.85 in the HCC-Genus dataset. The Adaboost method with a soft vote aggregator also performed well, achieving AUROC scores of up to 0.84 and performing equally as well as the PB-MVBoost method in some cases. In the IBD1 and Breast2 datasets a simple concatenation of modalities performed better than the multi-modal methods. However, further examination of the feature selection results in Section 3.4 will reveal that the features selected by this method are less stable than those selected by the multi-modal methods.

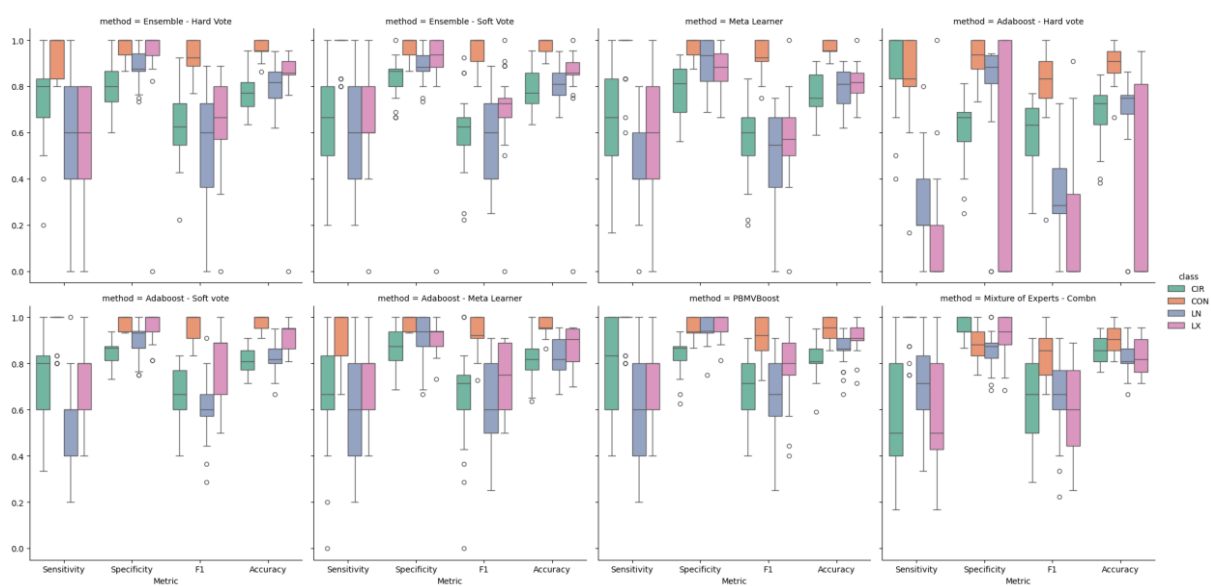

*Figure 4. Performance of the multi-modal models on the HCC (Genus) dataset. The colours of the bars represent the patient classes. Each group of boxplots in each subplot represents one metric, from left sensitivity, specificity, F1 score ad accuracy.*

A comparison of the performance of the best multi-modal method in each dataset to the best individual modality from the same dataset is shown in Table 4. In almost all cases the best multi-modal method outperforms the best individual modality. The exception is the Breast2 dataset where the performance is equal. This is evidence of the benefit of multi-modal data integration.

| Dataset | | Modality/Method | AUROC | F1 | Acc | Sens | Spec |
|---|---|---|---|---|---|---|---|
| **HCC - Genus** | Ind | CYT | 0.77* | 0.64 (±0.15) | 0.83 (±0.07) | 0.65 (±0.14) | 0.88 (±0.05) |
| | MM | PB-MVBoost | 0.85 | 0.77 (±0.11) | 0.89 (±0.06) | 0.77 (±0.15) | 0.93 (±0.06) |
| **HCC - Species** | Ind | CYT | 0.77* | 0.64 (±0.15) | 0.83 (±0.07) | 0.65 (±0.14) | 0.88 (±0.05) |
| | MM | PB-MVBoost | 0.84 | 0.75 (±0.13) | 0.88 (±0.06) | 0.76 (±0.16) | 0.92 (±0.06) |
| **IBD1** | Ind | MTB | 0.56 | 0.39 (±0.17) | 0.65 (±0.11) | 0.41 (±0.22) | 0.7 (±0.25) |
| | MM | Concatenation | 0.61 | 0.46 (±0.19) | 0.69 (±0.08) | 0.48 (±0.2) | 0.74 (±0.17) |
| **IBD2** | Ind | METAB | 0.76 | 0.68 (±0.06) | 0.79 (±0.07) | 0.69 (±0.07) | 0.83 (±0.07) |
| | MM | Adaboost - Soft | 0.8 | 0.74 (±0.05) | 0.82 (±0.06) | 0.74 (±0.05) | 0.86 (±0.06) |
| | | PB-MVBoost | 0.8 | 0.73 (±0.07) | 0.82 (±0.06) | 0.73 (±0.09) | 0.86 (±0.06) |
| **Breast1** | Ind | CLIN | 0.8 | 0.65 (±0.2) | 0.85 (±0.06) | 0.71 (±0.18) | 0.9 (±0.08) |
| | MM | Meta Leaner | 0.82 | 0.71 (±0.22) | 0.89 (±0.05) | 0.73 (±0.25) | 0.92 (±0.04) |
| **Breast2** | Ind | PROT | 0.74 | 0.57 (±0.36) | 0.91 (±0.04) | 0.58 (±0.38) | 0.93 (±0.07) |
| | MM | Concatenation | 0.74 | 0.58 (±0.37) | 0.92 (±0.06) | 0.59 (±0.4) | 0.93 (±0.1) |

### 3.3. Optimal Subset of Modalities

The incremental model, described in Section 2.5, was used to determine the most predictive subset of modalities in each dataset. For some datasets, namely IBD2, Breast1 and Breast2, a single modality gave the best predictive performance, while for others, namely HCC and IBD1, a small number of modalities gave equal or better performance than the full set of modalities. The results of this analysis are given in Table 5, which shows the degree to which performance improved as each modality was removed and the order in which the modalities were removed. The subset of modalities selected as optimal corresponds to the top performing individual modalities in the two datasets, confirming the validity of the incremental model.

| Dataset | Best Subset | Modality removed | Performance after removal |
|---|---|---|---|
| HCC | CLIN, CYT, METAB | None | 0.68 |
|  |  | OralSpecies | 0.70 |
|  |  | OralGenus | 0.71 |
|  |  | StoolSpecies | 0.72 |
|  |  | StoolGenus | 0.72 |
|  |  | Pathologic | 0.73 |
| IBD1 | MTB, MTG | None | 0.38 |
|  |  | VIR | 0.41 |
|  |  | MTX | 0.41 |
| IBD2 | METAB | None | 0.67 |
|  |  | CLIN | 0.69 |
|  |  | MICROB | 0.69 |
| Breast1 | CLIN | None |  |
|  |  | DNA | 0.22 |
|  |  | RNA | 0.25 |
|  |  | PATH | 0.25 |
| Breast2 | PROT | None | 0.44 |
|  |  | CLIN | 0.57 |
|  |  | MRNA | 0.58 |

*Table 5. Performance of the Incremental model in determining the best subset of modalities in each dataset, showing the degree to which performance improved as each modality was removed and the order in which the modalities were removed*

To further validate this method, Table 6 compares the AUC and F1 scores for each of the multi-modal methods on all modalities versus the optimal subset, for HCC and IBD1, as the optimal subset size for these two datasets was greater than one. It can be seen that each method performs as well on the optimal subset of modalities as it does on the full set of modalities.

| Dataset | Method | All modalities | | Optimal subset | |
|---|---|---|---|---|---|
| | | AUC | F1 | AUC | F1 |
| HCC | Concatenation + RF | 0.8 | 0.7 (±0.16) | | |
| | Voting: hard vote | 0.8 | 0.69 (±0.17) | 0.81 | 0.68 (±0.13) |
| | Voting: soft vote | 0.81 | 0.7 (±0.17) | 0.79 | 0.7 (±0.14) |
| | Meta Learner | 0.77 | 0.65 (±0.19) | | |
| | Adaboost: hard vote | 0.69 | 0.48 (±0.28) | 0.71 | 0.5 (±0.28) |
| | Adaboost: soft vote | 0.84 | 0.76 (±0.15) | 0.85 | 0.76 (±0.11) |
| | Adaboost: meta learner | 0.83 | 0.73 (±0.14) | 0.81 | 0.71 (±0.13) |
| | Mixture of Experts: combn | 0.71 | 0.51 (±0.28) | 0.74 | 0.6 (±0.19) |
| | Mixture of Experts: adaboost | | | | |
| | PB-MVBoost | 0.85 | 0.77 (±0.11) | 0.85 | 0.77 (±0.1) |
| IBD1 | Concatenation + RF | 0.61 | 0.46 (±0.19) | | |
| | Voting: hard vote | 0.56 | 0.39 (±0.25) | 0.55 | 0.37 (±0.24) |
| | Voting: soft vote | 0.57 | 0.42 (±0.23) | 0.59 | 0.45 (±0.17) |
| | Meta Learner | 0.5 | 0.3 (±0.27) | | |
| | Adaboost: hard vote | 0.54 | 0.33 (±0.27) | 0.51 | 0.18 (±0.2) |
| | Adaboost: soft vote | 0.56 | 0.4 (±0.22) | 0.6 | 0.45 (±0.19) |
| | Adaboost: meta learner | 0.53 | 0.31 (±0.26) | | |
| | Mixture of Experts: combn | 0.55 | 0.36 (±0.22) | 0.58 | 0.37 (±0.16) |
| | Mixture of Experts: adaboost | 0.56 | 0.33 (±0.25) | 0.56 | 0.33 (±0.16) |
| | PB-MVBoost | 0.57 | 0.39 (±0.22) | 0.56 | 0.39 (±0.19) |

*Table 4. Results showing the performance of each data integration method on the optimal subset of modalities in the HCC dataset. Figures in brackets are standard deviations. \* indicates statistically significantly worse performance than the best model.*

### 3.4. Feature Selection

As the primary purpose of developing these multi-modal models is to identify a clinical signature that can predict the development of the disease in question, the methods must also be judged on the clinical signature they identify. The desirable characteristics of a clinical signature include its length, its stability, the number of modalities from which it draws features and its accuracy in prediction. Ideally, a signature consisting of a smaller number of features and modalities will mean fewer tests for the patient and be more economical. The more stable, or reproducible, and the more accurate the feature selection results, the more confidence clinicians can have in their reliability in predicting disease.

The results of the feature selection for each multi-modal method applied to each dataset are summarised in Supplementary Table S3, which lists the number of features selected, the number of modalities those features are drawn from, the stability of the feature selection, measured using the relative weighted consistency index (Somol and Novovičová 2010), the predictive accuracy of the selected features and the mean of these two measurements. Note that as the Meta Learner uses a random forest as its meta classifier, and a random forest applies a non-zero feature importance score to every feature, rather than selecting a subset of features. Since every feature achieves a score on every iteration this method artificially achieves a perfect stability score of 1.

In every dataset, simple concatenation produced the least stable feature selection results. Concatenation of the modalities greatly increases the dimensionality of the data and feature selection is less stable in high dimensions. The data integration methods overcome this problem by training a separate classifier on each modality, illustrating another benefit of these methods.

In most datasets the PB-MVBoost method achieved the highest mean of stability and accuracy, but its signature length tended to be among the longest of all the methods and selected from the largest number of modalities. By contrast, in most cases the Adaboost model with a soft vote produced a slightly shorter signature with little loss in predictive accuracy or stability.

## 4. Discussion

In this work we have presented a benchmarking study of multi-modal multi-class machine learning techniques for late integration of multi-omics data, applied to five different datasets. We examined their predictive accuracy as well as the stability of the features they selected and compared the results of the multi-modal methods with those of the individual modalities. For each dataset, we also determined an optimal subset of modalities which performed as well as the full set, thus permitting patient diagnosis using fewer tests.

We employed existing and enhanced methods for late integration. Existing methods included a simple voting ensemble using hard and soft voting, a meta learner and the PB-MVBoost algorithm. Enhancements were made to the multi-modal Adaboost algorithm to improve its predictive accuracy and a novel application of the mixture of experts model was developed which builds an expert for each class and combines the results of these experts using a novel gating function.

Overall, the multi-modal methods showed superior performance to the individual modalities, with the PB-MVBoost and the Adaboost models being the most predictive. This shows that different modalities can provide complementary information about a patient's disease state, and integrating those modalities in a single model can improve predictive performance. Therefore, the use of these data integration techniques is recommended.

Boosting is an ensemble technique that trains a series of weak learners and combines them to form a stronger learner, improving predictive accuracy by reducing overfitting. Because of its ability to reduce overfitting, boosting is particularly suited to datasets that have a large number of features and a small number of samples, which is a characteristic of the data sets examined in this work. In addition, the two boosted methods, PB-MVBoost and Adaboost, calculate a weight for each modality, thereby prioritising the more predictive modalities. Combining boosting, modality weighting and data integration, gives these methods an advantage, so it is not surprising that they performed well.

In contrast, the meta learner, mixture of experts model and voting ensembles give equal weight to each modality, and our results on the individual modalities show that some modalities are more predictive than others. Therefore, the less predictive modalities are likely to be detrimental to the overall predictive power of the model.

Further, our results from the incremental model show that it is not always necessary to incorporate all modalities of data in a model to achieve the maximum predictive power. Finding the right subset of modalities to maximise predictive performance is crucial, not only to optimise the model outcome, but also for future clinical use, where a smaller set of modalities can simplify and reduce the cost of screening patients for a disease. Therefore, the use of our incremental model to determines the most predictive subset of modalities is recommended.

In the IBD1 and Breast2 datasets, a simple concatenation was the best performing model. The proteomic modality of the Breast 2 dataset significantly outperforms the other two modalities and is identified by the incremental model as being the best subset of modalities on its own. Therefore, it seems likely that the other two modalities do not contribute significantly to the model and that the proteomic modality dominates in the concatenation model.

In methods where the final decision could be made via a hard vote or soft vote, such as the voting ensembles and Adaboost, the soft vote was the better performer each time. Averaging the probability scores for each class in each modality, as in the soft vote, gives a finer tuning of the results than a simple hard cut-off, as in the case of the hard vote. In a hard vote ensemble, the more modalities in the model, the greater the chance of errors compounding and confusing the final decision.

The ability to identify a stable set of predictive features is an essential, yet often neglected, aspect of modelling any biological system with the aim of knowledge discovery, as these features are likely to provide valuable insights into the underlying biological processes. However, stability must be considered in conjunction with predictive performance. A method that selects the same set of features on each iteration will be very stable, but may have poor predictive accuracy, while another method that selects a different set of features on each iteration may be quite accurate but highly unstable and unsuitable for knowledge discovery. Our results show that the PB-MVBoost model and the Adaboost model produced the most accurate and stable feature selections, with the Adaboost model generally producing a slightly shorter clinical signature.

Based on our results, we can provide the following recommendations for integrating multi-modal, multi-class data. We recommend first examining the individual modalities to identify which are the most predictive. Following this the incremental method should be applied to determine whether an optimal subset of modalities can be found, and this optimal subset should align with the most predictive individual modalities. Finally, we recommend the training of a PB-MVBoost or Adaboost model with a soft vote for integrating the data modalities.

## 5. Conclusion

The ability to integrate data from multiple modalities can allow more accurate modelling of a patient's clinical outcome, as these modalities can provide complementary information. Such modelling may also assist in understanding the mechanisms underlying complex diseases and help identify novel biomarkers to aid in diagnosis.

The aim of this paper was to compare the performance of ensemble machine learning techniques capable of late integration of multi-class data from different modalities. The ensemble methods and their variations tested were i) a voting ensemble, with hard and soft vote, ii) a meta learner, iii) a multi-modal Adaboost model using a hard vote, a soft vote and a meta learner to integrate the modalities on each boosting round, the PB-MVBoost model and a novel application of a mixture of experts model. These were compared to simple concatenation as a baseline.

We examined the predictive accuracy of these methods, as well as the stability of the features they selected, and the size of the clinical signature determined, by applying them to an in-house dataset containing data collected from patients with liver disease (Behary et al. 2021), and validated them on data from four publicly available multi-omics datasets. Our results showed that two boosted methods, PB-MVBoost and Adaboost with a soft vote were the overall best performing models, both in terms of predictive accuracy and stability of feature selection. We also provided a means of determining an optimal subset of modalities that could lead to a smaller clinical signature without loss of predictive accuracy. Finally, we have provided recommendations for the integration of multi-modal multi-class data.

## Declarations

### Conflicts of interest/Competing interests
Author Fatemeh Vafaee is a member of the editorial board of Artificial Intelligence Review.

## Data Availability

The HCC dataset is available from the authors upon reasonable request. Data availability for the other datasets can be found in the relevant papers – references are given in Table 1.